# Exploring Neural Net Augmentation to BERT for Question Answering on SQUAD 2.0

**Eric Hulburd**
University of California, Berkeley
ehulburd@berkeley.edu

**Suhas Gupta**
University of California, Berkeley
suhas.gupta@berkeley.edu

## Abstract

Enhancing machine capabilities to answer questions has been a topic of considerable focus in recent years of NLP research. Language models like **E**mbeddings from **L**anguage **Mo**dels (ELMo)[1] and **B**idirectional **E**ncoder **R**epresentations from **T**ransformers (BERT) [2] have been very successful in developing general purpose language models that can be optimized for a large number of downstream language tasks. In this work, we focused on augmenting the pre-trained BERT language model with different output neural net architectures and compared their performance on question answering task posed by the **S**tanford **Q**uestion **A**nswering **D**ataset 2.0 (SQUAD 2.0) [3]. Additionally, we also fine-tuned the pre-trained BERT model parameters to demonstrate its effectiveness in adapting to specialized language tasks. Our best output network, is the contextualized CNN that performs on both the unanswerable and answerable question answering tasks with F1 scores of 75.32 and 64.85 respectively.

## 1. Introduction and dataset

There has been rapid progress made by researchers on the question answering task posed by SQUAD 2.0 but there remains ample opportunity to improve prediction of unanswerable questions. In this work, we use the pre-trained BERT language model and train the three output network architectures neural network architectures for the language task: 1) **basic convolution neural network 2) contextual convolution neural network, and 3) recurrent neural net with LSTM units.** We also compare training with fine tuning the BERT model to training these architectures on fixed features extracted from BERT. We used the SQUAD 2.0 question answering dataset in this work. SQUAD stands for **S**tanford **Q**uestion **A**nswering **D**ataset [3] and is a dataset with questions posed by crowdsourcing based on Wikipedia articles. This dataset poses a unique challenge to NLP algorithms by adding unanswerable questions to SQUAD v1.1[3]

## 2. Baseline performance

As of this writing, the best scores on the SQUAD 2.0 task are 86.8% exact match and 89.4 F1. The raw score for BERT posted on May 20, 2019 was 83.6% exact match and 86.0 F1. Using BERT base-uncased, the BERT run_squad.py script achieves 73.7% exact match and 77.0 F1. This latter result is what we hope to beat with model augmentation.

## 3. BERT baseline and fine-tuning

**B**idirectional **E**ncoder **R**epresentations from **T**ransformers is a language model that is pre-trained on a large corpus of text data to serve as a general purpose language model for downstream NLP tasks[2]. It uses the bidirectional training of transformer attention model [4] for application to language modeling. BERT uses the encoder part of the transformer architecture to create encodings that serve as a language model (Figure 1). Since the transformer model reads the entire sequence of words in both directions, BERT learns the contextual understanding between the word tokens.

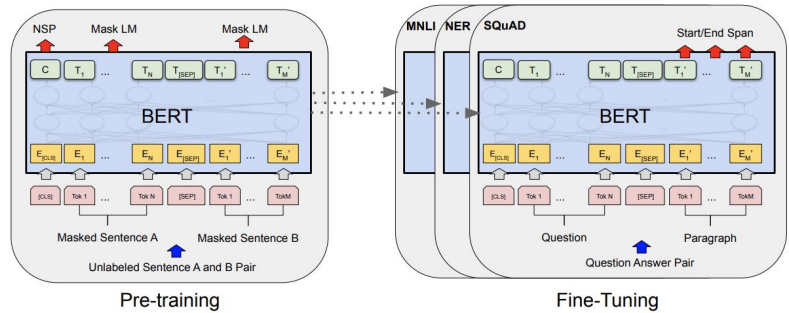

Figure 1 [2]. Figure from BERT authors representing the BERT pre-training and fine tuning flow and network architecture. We used the BERT pre-trained model for question answering on the SQUAD 2.0 database and fine-tuned the weights to improve model performance

We used the 'bert-base-uncased' pre-trained model in our work which consists of 12 transformer layers, 12 attention heads and outputs a word embedding matrix of (sequence length x 768) dimensions. We chose this model in order to limit training time and compute resource requirements, which was essential to explore the wide range of architectures and hyperparameters resulting from the proposed network.

## 4. Design of experiments

While the researchers of BERT have used the addition of a single affine layer to BERT embeddings to adapt the model to downstream NLP tasks, we explore the addition of complex neural network on top of the BERT embeddings. Our experiment design is shown in Figure 2, where we plug in different architectural layers on top of the BERT LM. The final logits, generated from a softmax layer, are for the start and end positions of the answer span within the reading comprehension in each training example.

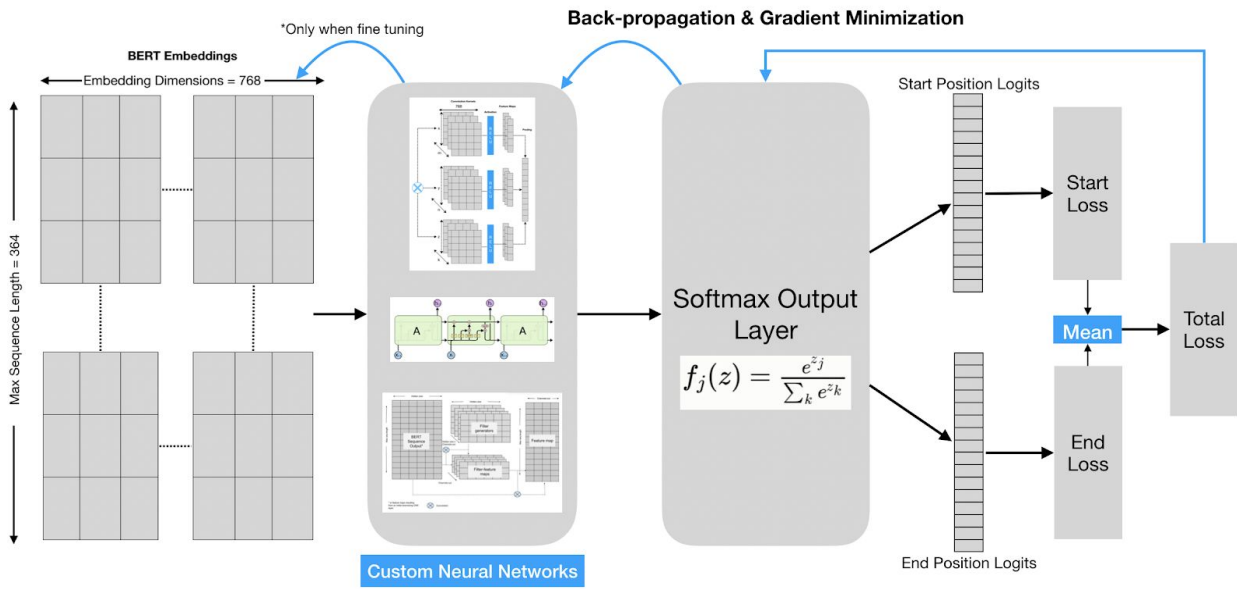

Figure 2. Experimental setup used in this work for training. The output from BERT language model was fed into our neural networks followed by a final softmax layer to generate two logit vectors: one each for start and end positions respectively. The total loss is the mean of the start and end position log loss values which is then minimized with the AdamOptimizer from the BERT repository. [All implementation was done Tensorflow 1.14]

## 5. Experimental output neural networks to augment BERT

### 5.1. Convolution neural network (1-D)

The first architecture we explored was a multi-channel convolution network that performs 1-D convolution on word embeddings as described in [5] and [6]. 1-D convolution uses the entire word embedding vector and scans through the vectors of different tokens based on the filter kernel size. Our hyperparameter search for tuning this model included several kernel sizes, number of filters per kernel, total filters per layer and total model layers. This CNN custom model performed worse compared to all other models (Table 1) due to loss of contextual information from BERT embeddings. This loss occurs because our search space for the correct hyperparameter combination is limited and it was computationally challenging to perform an exhaustive search. We used this model as baseline to demonstrate the need for generative filtering where the feature maps and other associated hyperparameters are learnt during the training phase. In terms of runtime, basic CNN was the quickest to train and experiment since convolution operations are easily parallelized on a GPU.

### 5.2. Recurrent neural net with LSTM units

Our second model was a recurrent neural network consisting of **L**ong **S**hort Term **M**emory (LSTM) cells. An LSTM cell is composed of three gates: a forget gate, input gate and output gate[7]. The forget gate is responsible for filtering out extraneous information from the previous token's output state. The input gate identifies which state values need to be updated as well as respective candidates for those values. Last, the output gate decides what should be passed on to the subsequent token's LSTM cell. We used Tensorflow's LSTMCell layer with static RNN module for our implementation [8].

Our intuition of applying LSTM to decode BERT embeddings for the SQUAD task was the ability of the LSTM cells to maintain an internal state representative of the question's significance, while outputting a state for each token corresponding to its relationship to that significance. We see in Table 2, that this model overfits the unanswerable questions while not learning to find tokens positions related to answerable questions.

### 5.3. Contextualized neural net with generative filter maps

We followed the architecture for a context aware convolutional neural network described in *Shen et al* [10] with some minor differences. The contextualized neural networks are initialized with a number of *meta-filters* or *filter generators* that are applied over the input sequences just as normal CNN filters. The resulting *filter-feature maps* are then themselves applied over the input sequences to produce a final feature map. Importantly, the *filter-feature maps* are unique to every input sequence of the dataset, whereas the *filter-generators* are a shared set of weights. Note, this requires *H* filter generators for every desired output channel of the final feature map, where *H* is the hidden size of the input sequence.

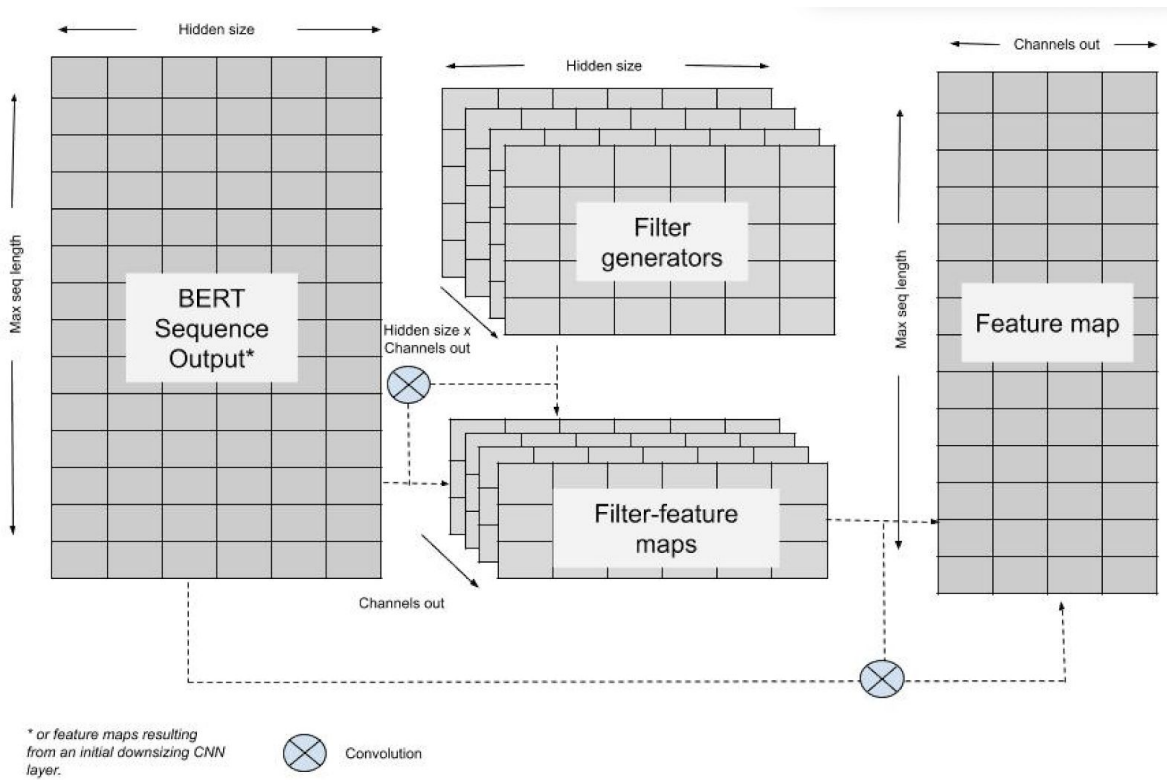

Figure 4. Contextual neural network using generative maps to find essential features in the word embeddings.

We can think of the *filter generators* as a set of filters that pull summary information from the entire input sequence relevant to the cost function - much as a human test-taker might do as they read a question along with input text. When these summaries, or *filter-feature maps*, are passed over the input sequences, they can recognize tokens in the original input sequences that are particularly relevant to the summary. Our approach differed from that of *Shen el al* [10] in that we follow the BERT model in concatenating questions and input text into our model, rather than use separate generator and convolutional modules as they did in their AdaQA model. Results in Table 2 illustrate the effectiveness of our approach with contextualized CNN.

## 6. Model training and evaluation

Model development was done using a training and evaluation heuristic where 10% of the SQUAD training data was reserved for evaluation. This produced 13,000 examples for evaluation and 117,000 examples for training. The batch size for one forward and backward pass was 32 when no BERT fine tuning was performed and 4 when BERT fine tuning was done (due to GPU memory limitations). We use the tensorboard API to write and visualize runtime summaries of scalars for accuracy and loss for the evaluation data. In order to prevent overfitting the model to the training data, we stopped the model training when the loss on the evaluation data reached an inflexion point.

$$N_{trainsteps} = (N_{samples} * N_{epochs})/N_{batchsize}$$
$$N_{evalsteps} = N_{trainsteps}/1000$$

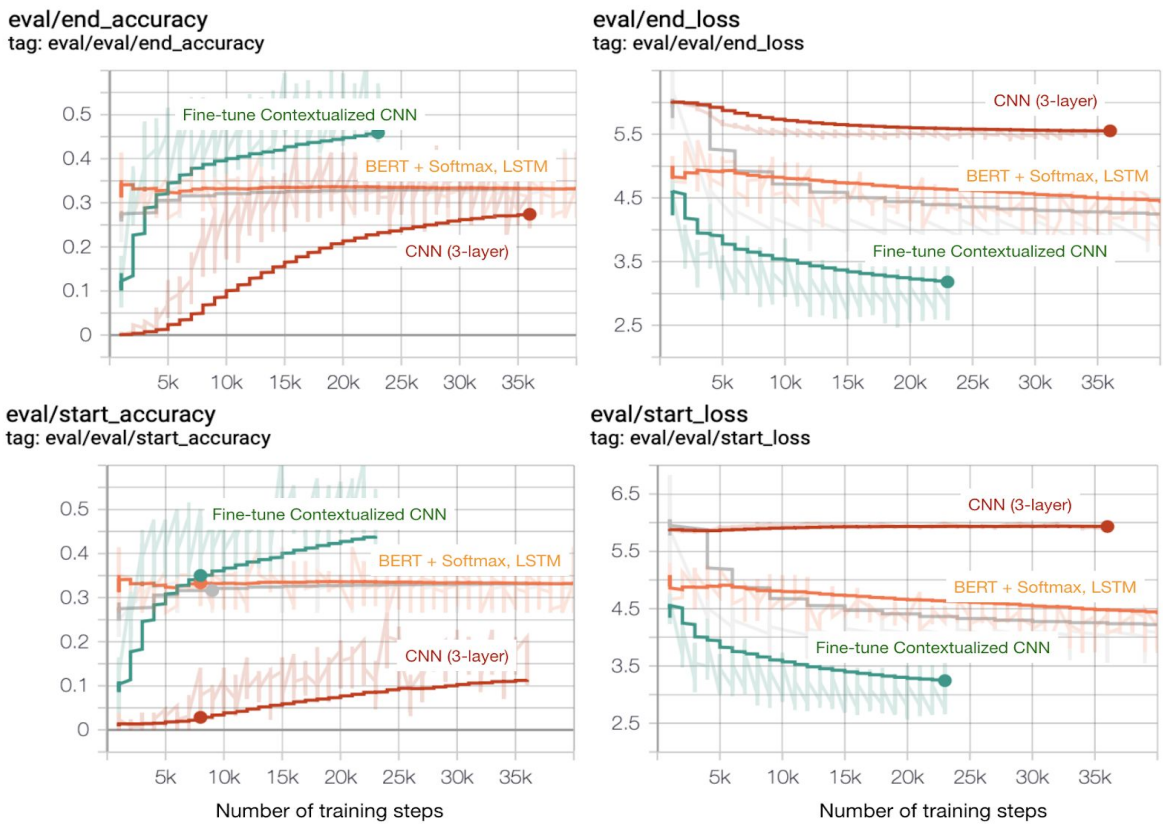

Figure 5. Training and evaluation flow used in this work. X-axis contains the number of training steps for each model (Cross validation was done once for every 1k steps of training forward & backward propagation). Generally, we tried to monitor the loss function as it progressed through the training loop and stopped the training when the loss reached an inflexion and began to increase. This occurred at different step size values for different neural architectures due to the training speed being different between them and the difference in batch size that could be utilized based on the number of parameters required to be stored in memory during training for each network.

## 7. Model evaluation

### 7.1. Fine tuning

We ran training and evaluation on all models with fine tuning and BERT extracted features (without fine tuning). Table 1 summarizes our results on the dev-2.0 SQUAD development dataset with the trained models.

|  | **Without fine tuning** | | **With fine tuning** | |
|---|---|---|---|---|
| **Model** | EM | F1 | EM | F1 |
| **BERT + fully connected** | 0.22 | 3.89 | 3.28 | 7.51 |
|  | 13.5 | 15.2 | 50.07 | 50.07 |
| **BERT + basic CNN** | 3.77 | 8.86 | 0.68 | 0.68 |
|  | 21.22 | 26.3 | 17.48 | 17.48 |
| **BERT + contextualized CNN** | 18.20 | 25.40 | 42.99 | 47.87 |
|  | 50.27 | 51.79 | **66.67** | **70.09** |
| **BERT + LSTM256** | 50.07 | 50.72 | 50.02 | 50.02 |
|  | 50.07 | 50.72 | 50.01 | 50.10 |

**Table1:** Model prediction performance with and without fine tuning the BERT base model parameters. EM= Exact Match and F1 = harmonic average of precision and recall. We observe that fine tuning BERT with augmented neural network improves the model performance in all cases. Contextualized CNN reaches the performance of original Squad implementation by Google [2]. The first number in each column is before threshold adjustment is made for null answer prediction and the second number is after the null threshold is adjusted and predictions are run again. There is substantial improvement in prediction score after thresholding is performed.

### 7.2. Observations
The following are interesting observations to note for our model performances as listed in Tables 1 and 2:
- The 1-D CNN network has a very low "no answer" score but a higher "has answer" score. Thus it tries to predict answers for most of the questions but doesn't learn any information about the impossible questions. We believe this is due to the loss of contextual information due to fixed filtering resulting in underfitting the evaluation data.
- The LSTM based model learns the task of impossible questions "too well", overfitting the "no answer"category. Since the dev-2.0 dataset from SQUAD 2.0 has a high fraction of impossible questions, we see from Table 2, that LSTM has very high scores for this question category.
- The contextualized CNN finds balanced medium between the two extremes of basic CNN and LSTM. Table 2 results show that generative filtering is able to learn creating the best kernel maps for the CNN to keep relevant contextual information from the BERT embeddings and adapting to SQUAD language task during the decoding stage. **This is thus our model of choice** in this work as it predicts both the "no-answer" and "has-answer" questions' answers with fairly good accuracy which is the challenge posed by SQUAD 2.0 language task.

|  | **Overall** | | **No Answer** | | **Has Answer** | |
|---|---|---|---|---|---|---|
| **Model** | EM | F1 | EM | F1 | EM | F1 |
| **BERT + fully connected** | 50.07 | 50.07 | 99.95 | 99.95 | 0.05 | 0.05 |
| **BERT + basic CNN** | 17.48 | 17.48 | 0.35 | 0.35 | 1.012 | 34.67 |
| **BERT + contextualized CNN** | **66.67** | **70.09** | **75.32** | **75.32** | **57.99** | **64.85** |
| **BERT + LSTM256** | 50.01 | 50.10 | 99.86 | 99.86 | 0.033 | 0.198 |

**Table1:** Values are shown for fine-tuned and null-threshold-adjusted predictions runs for all models. The basic CNN model underfits the eval data and tries to predict answers with less contextual information. The LSTM model overfits the no-answer data and predicts no-answer with high accuracy but doesn't predict answerable questions well. The contextualized CNN model is a generative model that learns the feature maps during the training and thus finds the right hyperparameter combination for predicting both the unanswerable and answerable questions with decent accuracy. Thus, this is the model of choice in our work as it handles the SQUAD 2.0 task better than other models.

## 8. Challenges and future work
In order to produce better scores we could have made two changes to our approach,:
1. We could have placed a higher priority on standardizing training time or stopping criteria. The major challenge we encountered was the unpredictable training time of different architectures and hardware. Due to resource constraints, we were forced to kill some training jobs prematurely.
2. Secondly, because we wanted to explore the differences between fine tuning and feature extraction, we departed from much of the BERT code base. While the BERT repository does include an *extract_features.py* script for saving sequence output to a *tf_record* file, we found it insufficient to support the additional processing required for the SQUAD 2.0 task. As a result, we lost performance parity between BERT repo's *run_squad.py* result and our own fine tuned fully connected network. In retrospect, it would have been worthwhile to amend the scripts provided by the BERT repository to run additional layers between the final BERT sequence output as well as more fluidly compare results between fine tuning and feature extraction approaches.

## 9. Conclusion

In this paper, we explored a variety of different architectures to augment the BERT architecture for use on the SQUAD 2.0 question answer task. We additionally explored the merits of training these augmented layers on extracted BERT features versus fine tuning the BERT layers for the SQUAD 2.0 task. Our fine tuning results unanimously outperformed the non-fine tuned trained models. However, because in-group rank between fine tuning and extracted features approaches were similar, it may make sense in some cases to use the extracted features approach to more quickly train over a wide array of architectures and hyperparameters and then train top candidates with the fine tuning approach.

It is evident from our results that we were not able to beat the performance of the BERT *'run_squad.py'* script using the same SQUAD 2.0 dataset and pre-trained BERT checkpoints. This calls for an additional attempt that more closely maintains training parity in order to conclusively assess the benefit of augmenting BERT with a contextualized CNN layer for the SQUAD 2.0 task. However, with respect to the explored architectures, our results support the merits of the contextualized CNN. It demonstrated the capability to detect both unanswerable and answerable questions with better scores than all other architectures implemented in this paper.